# 3D Face Parsing via Surface Parameterization and 2D Semantic Segmentation Network

Wenyuan Sun, Ping Zhou, Yangang Wang, *Member, IEEE*, Zongpu Yu, Jing Jin, Guangquan Zhou

*Abstract*—Face parsing assigns pixel-wise semantic labels as the face representation for computers, which is the fundamental part of many advanced face technologies. Compared with 2D face parsing, 3D face parsing shows more potential to achieve better performance and further application, but it is still challenging due to 3D mesh data computation. Recent works introduced different methods for 3D surface segmentation, while the performance is still limited. In this paper, we propose a method based on the "3D-2D-3D" strategy to accomplish 3D face parsing. The topological disk-like 2D face image containing spatial and textural information is transformed from the sampled 3D face data through the face parameterization algorithm, and a specific 2D network called CPFNet is proposed to achieve the semantic segmentation of the 2D parameterized face data with multi-scale technologies and feature aggregation. The 2D semantic result is then inversely re-mapped to 3D face data, which finally achieves the 3D face parsing. Experimental results show that both CPFNet and the "3D-2D-3D" strategy accomplish high-quality 3D face parsing and outperform state-of-the-art 2D networks as well as 3D methods in both qualitative and quantitative comparisons.

## I. Introduction

WITH the development of computer vision, face technologies have become increasingly popular due to the abundant achievements and the wide application in the last decades. Among the studies, face parsing assigns pixel-wise semantic labels as face representation for computers, and it has attracted more and more attention because of its fundamental support to other advanced face technologies, such as facial expression transfer [1], face editing [2], and other face information analysis [3]. Some face image processing methods, such as illumination transformation [4] and deblurring [5], as well as face editing with Generative Adversarial Network (GAN) [6], are also based on face parsing as an important step.

In recent years, the Convolutional Neural Network (CNN) has achieved significant success in many applications [7], [8], whose performance is far superior to that of traditional methods such as prior-based algorithms [9]. To boost the performance of face parsing further, well-designed computation operations, convolution pathways, network architectures, and specific algorithms with different strategies have been presented [10]-[12]. Based on these methods, new applications are achievable, such as advanced face parsing via end-to-end hybrid neural network to assist traditional Chinese medicine inspection [13].

However, to the best of our knowledge, most face parsing works still focused on 2D images at present, and the 3D face parsing is considerably limited as a grand challenge compared with the 2D image-based works. Although great achievements have been made on 2D face parsing in the past, there are still limits for these methods due to the 2D imaging mechanism. Two main limits should be pointed out in 2D face parsing: Firstly, the performance may be greatly affected by the environment during the 2D image sampling, especially illumination variations. Bad illumination such as over-exposure and under-exposure, may lead to severe loss of textural information, causing underrepresented prediction for most existing benchmarks due to their high sensitivity to image sampling conditions, as shown in Fig.01 (a), (b) and (c). Secondly, the performance of 2D face parsing is sensitive to occlusions caused by different poses, because some important information may be incomplete in specific viewpoints, as shown in Fig.01 (d). The critical reason for these two limits in 2D image-based face parsing is that such pixel-wise semantic labeling is highly texture dependent, while it is irrespective of 3D spatial structure information. Some image-based methods [14], [15] have been proposed to solve these problems by regressing coefficients for 3D information, while the effectiveness and robustness need improvement.

With the rapid development of 3D measurement, accurate 3D face data is now available based on stereo vision, structured light, light field, and ToF (time of flight) methods [16], [17],

W. Sun was with School of Biological Sciences & Medical Engineering, Southeast University, 210096 Nanjing, P.R. China. (e-mail: wenyuansun1998@gmail.com).

P. Zhou is the corresponding author. He is with School of Biological Sciences & Medical Engineering, Southeast University, 210096 Nanjing, P.R. China (e-mail: capzhou@163.com).

Z. Yu was with School of Biological Sciences & Medical Engineering, Southeast University, 210096 Nanjing, P.R. China (e-mail: weensom@163.com).

Y. Wang is with the School of Automation, Southeast University, Nanjing 210096, China, and also with the Shenzhen Research Institute, Southeast University, Shenzhen 518063, China (e-mail: ygwangthu@gmail.com).

J. Jin is with City University of Hong Kong, 210096 Nanjing, P.R. China (e-mail: jingjin25-c@my.cityu.edu.hk).

G. Zhou is the corresponding author. He is with School of Biological Sciences & Medical Engineering, Southeast University, 210096 Nanjing, P.R. China e-mail: guangquan.zhou@seu.edu.cn).

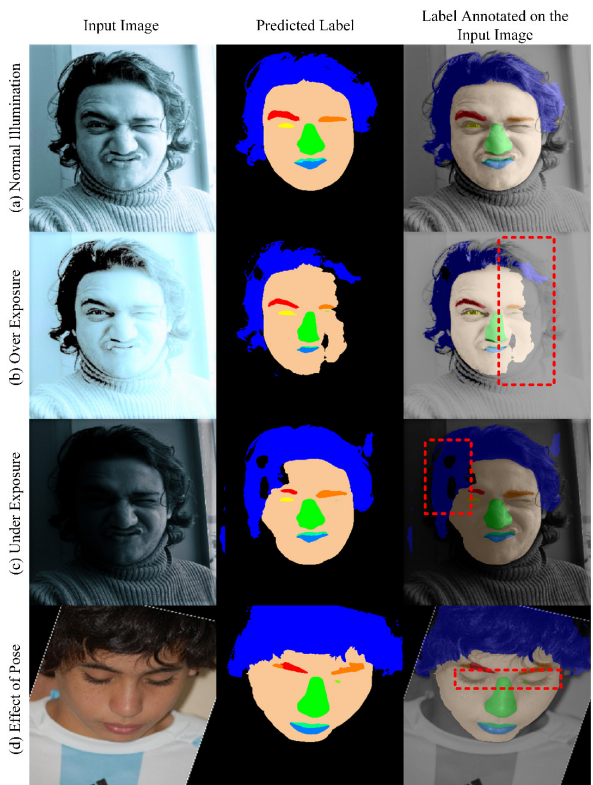

Fig. 1. Face parsing results under different illumination variations and poses, with 2D semantic segmentation network trained on Helen database

which make it possible for 3D face parsing. Unfortunately, although many 3D data segmentation works have been presented, studies of 3D face parsing are still limited. Generally, the 3D data is represented in three types: voxel data, point cloud, and mesh data. In the case of voxel-based methods [18], [19], the 3D data is projected on 3D grids first to obtain the regular structure for easier computation, while it results in low flexibility and efficiency. In contrast, point cloud and mesh data are more suitable for surface representation of the 3D face data, while their lack of regular data structure increases the difficulty of computation and synthesis. Recently, advanced methods for surface segmentation have been proposed in both point cloud networks [20]-[22] and graph-based deep learning technologies [23], [24]. However, these works mainly focused on scene segmentation, such as indoor scenes, and the performance is limited in face parsing. In addition, some 2D image-based methods, including multi-view methods [25], [26] and RGB-D methods [27], [28] have been proposed for the 3D data segmentation. However, for face parsing, these methods still suffer from occlusion because of their 2D sampling mechanism, and the multi-view methods result in great computation consumption.

In this paper, we propose a 3D face parsing method via the "3D-2D-3D" strategy for 3D mesh data. Compared with multi-view 2D image-based methods, the simply-connected open 3D face surface sampled in the wild is transformed to a topological disk-like 2D face image via the face parameterization algorithm in our method, where the noise is filtered and conformal mapping is used to preserve the texture information of 3D face surface. Although the 2D mapped face image is disk-like and distorted, it maintains the geometric information on the original structures. Subsequently, a specific 2D convolutional neural network is designed to accomplish 2D face parsing for 2D conformal parameterized face images, which is called Conformal Parameterization Face-parsing Network (CPFNet) in this paper. For the distortion due to the conformal parameterization, the CPFNet captures multi-scale structural information and aggregates hierarchical features, improving the performance with affordable computation consumption. At last, 2D parsing results from CPFNet are re-mapped to 3D face data, so that the 3D face parsing is accomplished. Experiments have been carried out to compare with other state-of-the-art methods, and our method outperforms others referring to the results.

The main contributions of this paper include:

(1) A "3D-2D-3D" strategy was proposed to accomplish 3D face parsing. The 3D face data is transformed to only one disk-like 2D image and the facial component labeling is achieved with a 2D semantic segmentation network. The semantic labels are re-mapped back to the 3D face data to achieve 3D face parsing. To the best of our knowledge, no work has been reported on employing the 3D face parsing through a 2D segmentation network using one 2D face image.

(2) The 3D face data is mapped to a 2D domain via the face parameterization algorithm. The noise is reduced during the algorithm first and the conformal parameterization is achieved efficiently. No additional occlusions occur during the mapping and the 3D geometric information is maintained completely.

(3) According to the characteristics of the 2D mapped face image, a 2D segmentation network (CPFNet) is proposed with multi-scale analysis and feature aggregation. Compared with state-of-the-art networks, our method is more appropriate to the disk-like parameterized face image with distortion.

The remainder of this paper is organized as follows. Section II summarizes related works on conformal parameterization, face parsing, and 3D surface segmentation. In Section III, we introduce our method in detail, including the face parameterization algorithm and our CPFNet. Experimental results and comparisons with other methods are presented in Section IV. Finally, this paper is concluded in Section V.

## II. RELATED WORKS

### A. Conformal Parameterization

In our "3D-2D-3D" strategy, the surface parameterization algorithm is used to map the 3D spatial surface to the 2D space based on some specific principles. Among all methods, the conformal mapping method shows remarkable potential in applications due to the advantage of preserving angles between any two lines on the surface, involving feature extraction and information analysis. In 2003, a systematic approach for conformal structure computation of general 2D surfaces is proposed [29], providing the first practical algorithm to compute conformal structures for closed meshes. Later, Gu *et al* [30] proposed a general method to achieve the conformal parameterization based on cohomology group structure and applied the method in the brain surface mapping. Besides, other computation approaches such as [31], [32] provide 3D data researchers with more fundamrntal algorithm, promoting the development of 3D surface analysis in a variety of fields such as brain research [33], [34].

The 3D face data we focus on is the simply-connected open surface, and the boundary of the mapped 2D face image is fixed. Compared with the method for free boundary mapping such as [31], the presence of the fixed boundary results in more complexity, and the harmonic parameterization induces conformality distortions due to fixed boundary. To solve such problems, some methods achieved mapping based on theories such as Ricci Flow [35] and Yamabe Riemann map [36], where the distortion is limited in an acceptable range. To improve the computational efficiency, in 2015, Choi *et al* [37] presented a scheme for fast disk conformal parameterization. Later, he presented a new conformal parameterization algorithm for multiply connected surfaces [38].

### B. Face Parsing

In this work, face parsing assigns every pixel, vertex, or triangle patch with a categorical semantic label, providing an accurate region representation of the face. Before the achievement of deep learning, face parsing is mainly based on methods such as prior knowledge [9]. Besides, exemplar is also a popular tool, such as [39], where facial components are labeled through aligned exemplar transformation.

In early works of face parsing via deep learning, the facial component labeling such as [40], [41] is achieved by the simple neural network models assisted with other complicated algorithms. In 2015, Long *et al* [42] proposed the fully convolutional network with encoder-decoder architecture, providing a network structure to achieve region segmentation. In 2016, a face parsing algorithm based on cascaded CNN was proposed by Jackson *et al* [7] based on landmark-guided strategy. To improve the performance, advanced networks with new structures were presented, such as Guo's network [43] with residual strategy to extract and decode features, the adoptive receptive field technology by Wei *et al* [10], and EHANet by Luo *et al* [11] with stage contextual attention mechanism and semantic gap compensation block. These works improve face parsing performance and ensure high-quality hierarchical analysis. Except for the network structures, some ingenious algorithms are also presented for practical problems, such as ROI Tanh-wrapping proposed in [12]. Recently, the adaptive graph representation method [44] was introduced to accomplish accurate 2D face parsing by learning and reasoning for component-wise relationship exploitation. Besides, domain adaptation has also been employed in face parsing for specific images [45].

However, to the best of our knowledge, few works have been proposed to deal with 3D face parsing. Most studies still focus on 2D image-based facial component segmentation via deep convolution neural networks of 2D images.

### C. 3D Surface Segmentation

3D surface segmentation has become popular because of the increasing applications in 3D data. Generally, 3D surface segmentation can be accomplished by voxelization, 2D projection, point cloud, and graph-based methods, *et al*.

Voxelization methods construct the regular grid for 3D surface by projecting on 3D voxel grids [18], [19], but it occupies much memory and consumes much computation resource. To solve this problem, low-resolution voxel grid and octree-based techniques are employed respectively, while the accuracy becomes lower.

Multi-view method is often applied in 3D segmentation using a set of 2D images sampled from several perspectives [25], [26], but such methods suffer from inflexibility due to its mechanism. In [46], a tangent convolution method is presented, projecting local neighborhood to local tangent planes. However, this method relies on tangent estimation heavily. RGB-D is also a 2D projection method aggregating RGB texture with spatial information [27], [28], but it still suffers from occlusion.

Point cloud and graph-based methods are often combined for sufficient features extraction. Pointwise MLP networks such as PointNet [20] and PointNet++ [21] uses methods including shared MLP, global max-pooling, T-Net as well as sampling and grouping layers to solve the feature transformation and contextual scale problems. Besides, point convolution methods such as KP-conv [22] and PointCNN [47] improve the local feature capturing of 3D geometry. Graph convolution methods alleviate the synthesis of local neighborhood information, such as the EdgeConv applied in DGCNN [23]. Mesh networks such as MeshCNN [48] utilize mesh data directly to analyze 3D shapes, it is hard to apply in 3D face data due to the challenge of point cloud resampling and robust 3D mesh reconstruction.

In addition, other methods such as manifold technologies also provide interesting ideas to analyze 3D surface, while most researches are focused on shape analysis [49] and some other tasks [50] so far. The manifold algorithm and model for 3D face semantic segmentation are limited.

### III. METHOD

The pipeline of our 3D face parsing method based on the "3D-2D-3D" strategy is shown in Fig.2. The 3D face data is transformed to the 2D disk-like face image via the face parameterization algorithm, and it is labeled via a semantic segmentation network called Conformal Parameterization Face-parsing Network (CPFNet), subsequently. At last, the 2D segmentation result is re-mapped to accomplish the 3D face parsing. In this section, we illustrate our method in detail.

### A. 3D Face Parameterization

To transform the sampled 3D face data to the high fidelity 2D image for the following 2D semantic segmentation, a 3D face parameterization method with two steps is proposed in this paper, as shown in Fig. 3. Due to the sampled noise in the raw data, an iteration denoising algorithm via local plane regression is employed first. Then, conformal parameterization is applied because of the feature preserving, followed by image rotation to generate high-quality data for the network. The two steps of this parameterization method are introduced in detail in the following sub-sections.

### 1) 3D Face Data Denoising

Because of the noise, coordinates of some vertices deviate in the direction in which the 3D human face data faces. Such deviation, which mainly distributes in regions of eyes and eyebrows, often causes random peaks and valleys on the 3D spatial surface. We propose an iteration denoising algorithm for vertex coordinate estimation and 3D face data optimization.

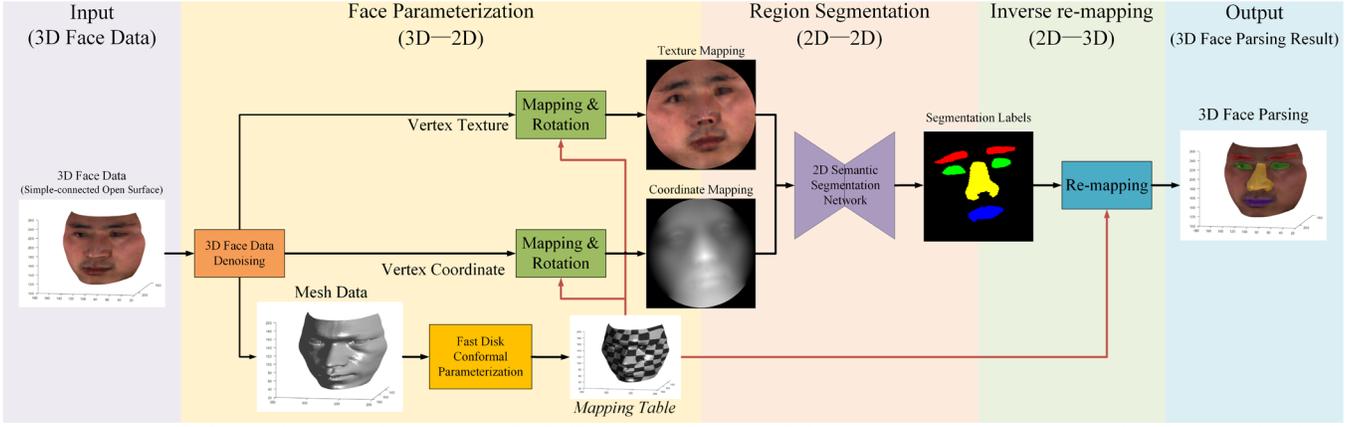

Fig. 2. Our proposed 3D face parsing method based on the "3D-2D-3D" strategy. The input 3D face data (simply-connected open surface) is mapped into 2D face image by the Face Parameterization, and the 2D face image is labeled by the 2D semantic segmentation network. After the 2D labeling, the result is re-mapped to 3D face data via the mapping table obtained in the Fast Disk Conformal Parameterization.

Three-dimensional vector $v = (x, y, z)$ represents the coordinate information of a vertex and vector $c = (r, g, b)$ for texture, where axis $Z$ represents the facing direction of the 3D face. Some vertices disturbed by the noise severely are far from their proper position near the surface, which are easy to be detected. Therefore, the denoising process for such vertices should be launched first based on the spatial dependence between vertex $v$ and its adjacency vertices, whose coordinates are $\{\hat{v}_1, \hat{v}_2, ..., \hat{v}_N\}$ and texture vectors are $\{\hat{c}_1, \hat{c}_2, ..., \hat{c}_N\}$. If there are more than $\alpha_A N$ adjacency vertices that satisfy $\|v - \hat{v}_i\|_2 > \epsilon$, the vertex $v$ is regarded as a vertex with severe noise, where $\alpha_A$ is a proportion parameter and $\epsilon$ is the threshold. The coordinate and texture will be replaced by:

$$\tilde{v} = \frac{1}{N}\sum_{i=1}^{N}\hat{v}_i, \tilde{c} = \frac{1}{N}\sum_{i=1}^{N}\hat{c}_i \quad (1)$$

In this case, sharp peaks and valleys caused by vertices far from the surface can be corrected based on their adjacency vertices. Although such threshold-based correction cannot reduce the noise sufficiently, correction of severe noise can still benefit the following denoising.

To correct the coordinate errors of the vertices left, whose noise intensity is lower than the threshold, an iteration surface optimization via local plane regression and weighted coordinate estimation is employed. Noted that the ideal 3D face data can be generally regarded as a smooth surface lack of local fluctuation, local area of vertex $v$ can be simplified as a plane $p$ in a small neighborhood $\sigma$. Let $k = (k_1, k_2, ..., k_n)$ to be different numbers of vertices that are involved in the following computation in $\sigma$, and $S_{(k_i)} = \{v_1, v_2, ..., v_{k_i}\}$ represents the corresponding $k_i$-closest vertex set. For each $k_i$ in $k$, a local plane $P_{(k_i)}$ attached with its Mean Squared Error ($MSE_{(k_i)}$) can be obtained via plane regression based on Least Squares Method, and the corrected $Z$ axis coordinate, $\tilde{z}_{(k_i)}$, is solved based on the $(x, y)$ and regressed plane $P_{(k_i)}$.

For each $k_i$ in $k$, a corrected coordinate $\tilde{v}_{(k_i)} = (x, y, \tilde{z}_{(k_i)})$ is obtained. For weighted estimation, the corrected vertex matrix $V_{(v)}$ is defined as:

$$V_{(v)} = \begin{pmatrix} x & y & \tilde{z}_{(k_1)} \\ & \vdots & \\ x & y & \tilde{z}_{(k_n)} \end{pmatrix} \quad (2)$$

Two types of weight are used during this optimization step for 3D face data denoising. To measure the confidence of the regressed local plane $P_{(k_i)}$, $MSE_{(k_i)}$ is used to compose the error weight matrix $W_{MSE}$, given by:

$$W_{MSE} = \left(\frac{1}{MSE_{(k_1)}} \quad \cdots \quad \frac{1}{MSE_{(k_n)}}\right)^T \quad (3)$$

Meanwhile, noted that ideal neighbor $\sigma$ should be as small as possible to be approximate to a plane, the maximum distance between vertices in $S_{(k_i)}$ and $v$, $d_{(k_i)}^{max}$, is applied in distance weight matrix $W_d$, given by:

$$W_d = \left(\frac{1}{\left(d_{(k_1)}^{max}\right)^2} \quad \cdots \quad \frac{1}{\left(d_{(k_n)}^{max}\right)^2}\right)^T \quad (4)$$

The correction weight matrix $W$ is calculated by:

$$W = \frac{W_{MSE} D W_d}{\sum_{i=1}^{n}\frac{1}{MSE_{(k_i)} \times \left(d_{(k_i)}^{max}\right)^2}} \quad (5)$$

where $D$ is a $n \times n$ diagonal matrix. With the corrected vertex matrix $V_{(v)}$ and correction weight matrix $W$, the corrected coordinate $\tilde{v}$ is estimated by:

$$\tilde{v} = W^T V_{(v)} \quad (6)$$

To improve the denoising and sufficiently correct the errors, the result $\tilde{v}$ of Eq. (6) is only the result of one iteration. Based on the upgraded 3D face data after the first-iteration correction, the iteration correction step is applied. The iteration will stop when the Mean Correction Step ($MCS$) is lower than the iteration threshold $\mu$, where the expression of $MCS$ after the $i^{th}$ iteration is given by:

$$MCS = \frac{1}{N_v}\sum_{j=1}^{N_v}\|v_j^{(i)} - v_j^{(i-1)}\|_2 \quad (7)$$

where $N_v$ is the number of vertices in the 3D face data.

*2) Conformal Parameterization*

Among various systematic computing approaches, the Fast Disk Conformal Parameterization (FDCP) proposed by Choi *et al* [37] is applied for efficient conformal parameterization of the simply-connected open surface, which is appropriate to our 3D face parsing work. In FDCP, the boundary conformality distortion correction cost most of the computation time, while regions near the boundary of 3D face data which are involved in such step are not important for our face parsing method, and its effect on the inner part of the mapped image from the 3D face data is considerably small. Therefore, the parameterization we apply in our study only contains the first two parts: the discrete harmonic mapping method for initial parameterization and the conformality improvement of the inner part via Cayley transformation and quasi-conformal theories.

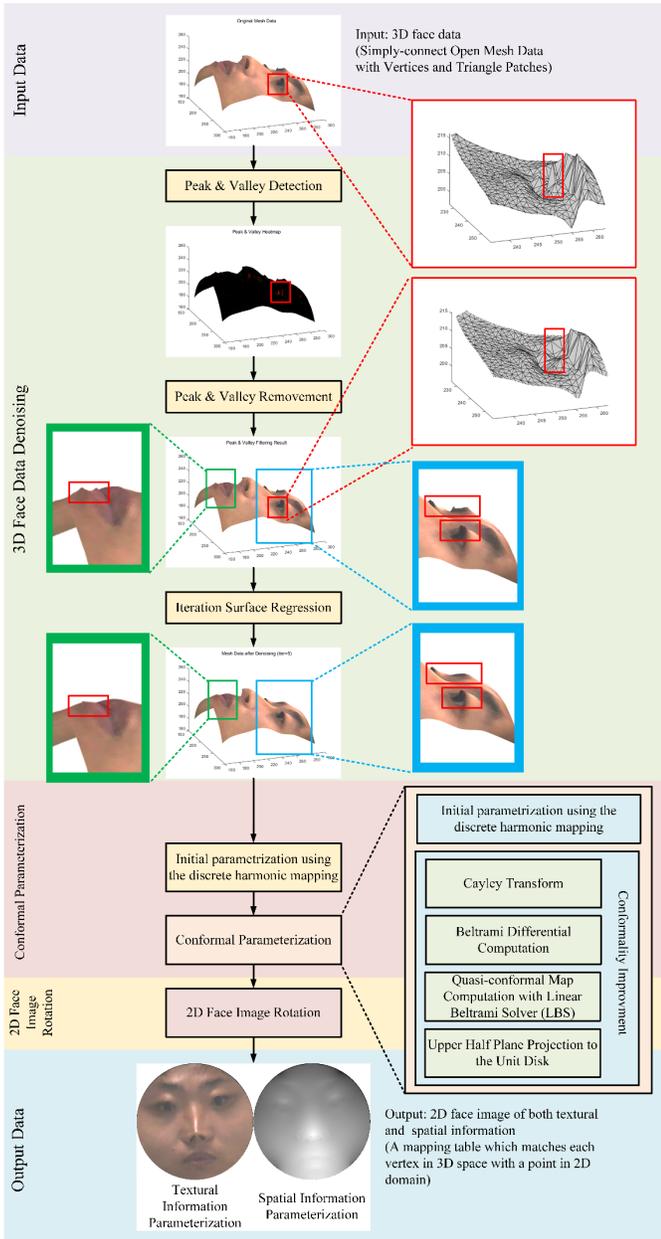

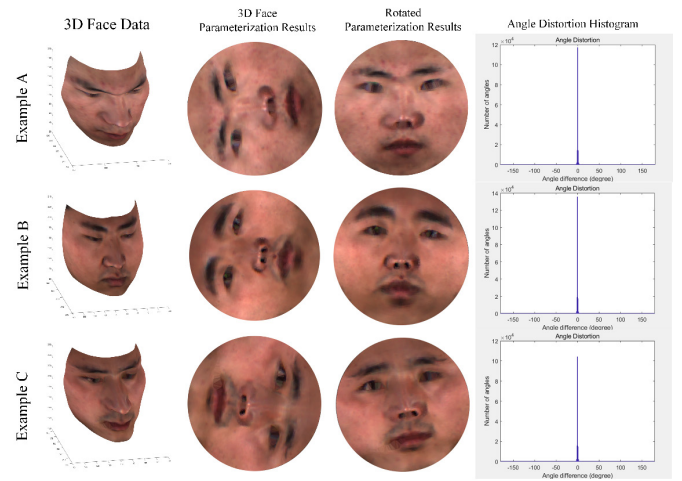

Fig. 4. Several results of our 3D Face Parameterization applied on 3D face data in BJTU-3D Dataset. 3D face data is transformed into 2D images and the conformality is evaluated by the angel distortion histogram.

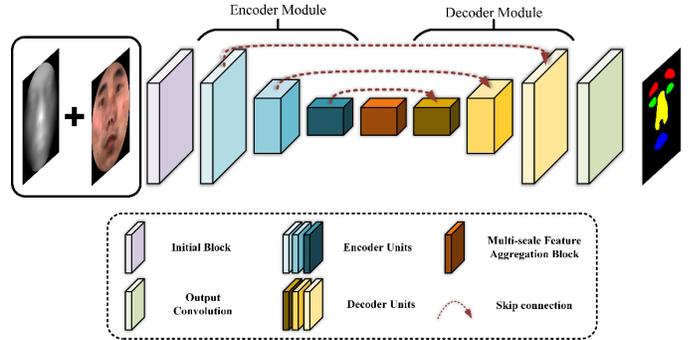

Fig. 5. The structure of our CPFNet for the semantic segmentation of facial components in conformal mapped face images. Basic elements including Initial Block, Encoder Unit, Decoder Unit, Multi-scale Feature Aggregation Block, and Output Convolution, construct the network with the Encoder-Decoder architecture and skip connections.

Fig. 3. The pipeline of our proposed 3D Face Parameterization in our "3D-2D-3D" Face Parsing method to transform 3D face data to 2D face images. 3D Face Data Denoising is firstly introduced in this algorithm, and Conformal Parameterization and 2D Face Image Rotation is applied. Parameterization result of both textural and spatial information is obtained, which can be fed in the 2D semantic segmentation network for face parsing.

As shown in Fig.4, 3D face data is mapped into 2D images with little conformality distortion referring to the angle distortion histogram. Each vertex $v$ with coordinate $(x, y, z)$ in 3D space $\mathbb{R}^3$ is matched with a point in a 2D disk domain $\mathbb{D}^2$ with the coordinate of $(u, v)$. By this mapping table, we can generate the 2D image by interpolation which contains both textural and spatial information. In our study, all 3D data faces Z axis direction, so the main spatial information is represented by normalized Z coordinate to record the depth.

As shown in Fig.4, most 2D mapped faces are rotated to undesired positions due to conformal parameterization. According to Gong's work about classification based on CNN and Mop-CNN, the prediction difficulty increases as rotation degree increases [51]. Therefore, the mapped face image after conformal parameterization should be approximately rotated to a proper position before being fed to the segmentation network.

*B. CPFNet*

After parameterization, the distorted, disk-liked 2D face image maintains the morphological feature of the human face, so they are still appropriate for face parsing. Whereas, compared with the other face parsing work with normal 2D face images, some particular features of the 2D conformal mapped face image inevitably increase the face parsing difficulty. The features mainly include: (1) Translation and deformation: During the conformal parameterization, the restrictive circular boundary constraints may cause translation and deformation of facial components, which is the primary challenge for 2D mapped face parsing. (2) Non-uniform sampling: Both vertices on the 3D mesh data and their mapped positions in the 2D domain distribute non-uniformly. Therefore, a network with multi-scale analysis and feature aggregation may show better performance in this situation. Besides, compared with 2D image datasets, the number of subjects in most 3D face datasets is limited even after data augmentation. So, it is critical to design an efficient and lightweight network for easy training.

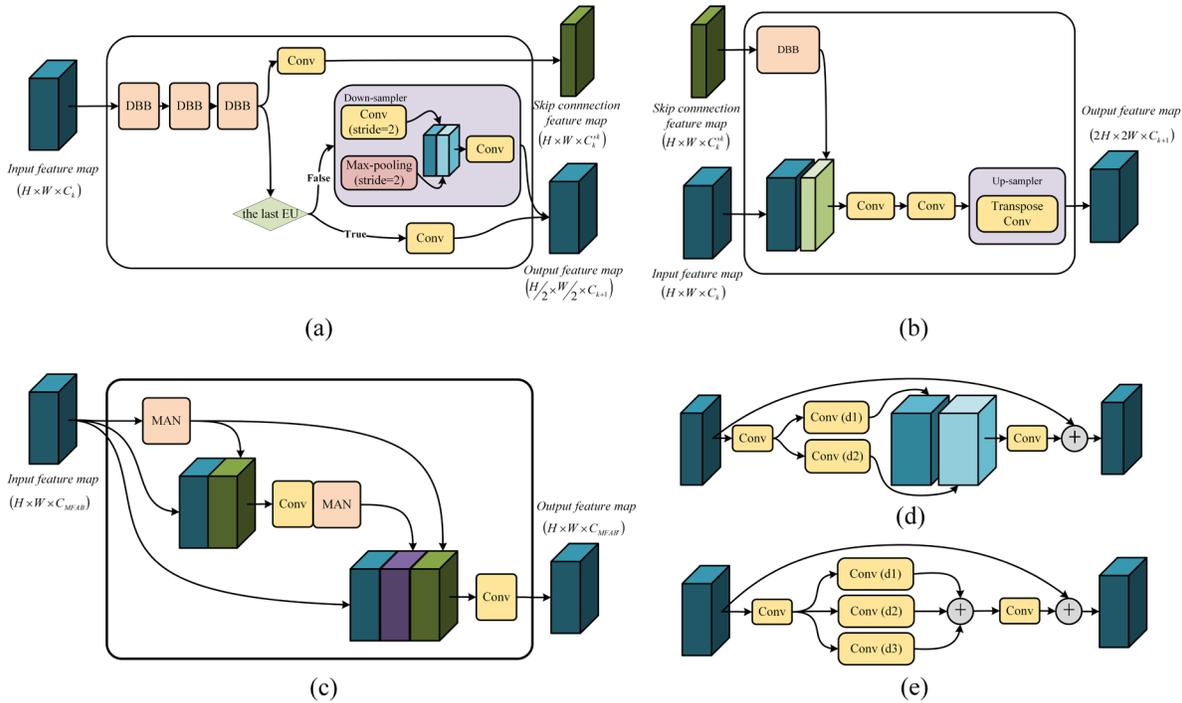

Fig. 6. The detailed structure of key elements of our CPFNet. (a) Encoder Unit (EU) for feature extraction. (b) Decoder Unit (DU) for semantic information reconstruction. (c) Multi-scale Feature Aggregation Block (MFAB) for feature fusion and hierarchical information analysis with its dense structure. (d) Dual-scale Bottleneck Block (DBB) as a critical block in EU and DU, where dual-scale features are exploited by the dual-convolution with different dilation rates. (e) Multi-scale Aggregation Node (MAN) as the basic block of MFAB, which combines multi-level multi-scale features with abundant semantic information.

In this paper, we proposed a 2D segmentation network called Conformal Parameterization Face-parsing Network (CPFNet) to accomplish the 3D face parsing. As shown in Fig.4, the input is a 2D face image with both texture and depth information generated from the face parameterization algorithm. The CPFNet inherits the common framework of encoder-decoder network for semantic segmentation. After the Initial Block decomposes the 4-channel input into basic feature maps, the encoder and decoder modules, respectively composed by cascaded Encoder Units (EU) and Decoder Units (DU), extract and integrate semantic information in the mapped face image delicately. Skip connections are applied between EUs and DUs, combining different feature maps of the same resolution. At the end of the decoder module, an output convolution generates the label map as the parsing result. Between encoder and decoder modules, the feature map has lower spatial resolution but more channels. Multi-scale Feature Aggregation Block (MFAB) is designed for sufficient information fusion and synthesis. The detailed structure of all modules is illustrated as follows.

*1) Encoder and Decoder Units*

Our proposed EUs and DUs are designed to be adequate for comprehensive information collection and multi-scale feature analysis. As shown in Fig. 6 (a) and (b), a convolutional block called Dual-scale Bottleneck Block (DBB) is presented and applied in both the EUs and DUs to extract feature and reconstruct information. As shown in Fig.6 (d), the DBB inherits the fundamental idea of the standard bottleneck block. To capture the multi-scale information with different receptive fields, the second convolution layer in the standard bottleneck block is replaced by two parallel convolutional pathways, whose filter sizes are the same while dilation rates are different. Subsequently, the feature maps from the parallel convolutional pathways are concatenated as a feature fusion operation for synchronous multi-scale feature analysis. With the short-cut structure, such operation can converge better during training.

In the EU, several cascaded DBBs (three as shown in Fig.6 (a) are applied for discriminative features extraction and adequate analysis of relationships between features of different scales. To provide the feature map with high resolution to the corresponding DU and generate feature map for the following EU, the cascaded DBBs are attached with two pathways, as shown in Fig.6 (a). One is a convolution layer to provide the feature map for skip connection, the other is an output module to produce the input feature map of the next EU or the following MFAB. A down-sampler is applied for EU, while a convolution layer is employed for MFAB. To gather context information, the down-sampler here combines both Max-pooling and convolution with strides, and the combined feature map is further analyzed with a convolutional operation, providing the output feature map with the resolution of $\left(\frac{H}{2} \times \frac{W}{2} \times C_{k+1}\right)$.

To assist the decoding computation, high-resolution feature maps are provided from EU and are introduced to corresponding DUs for feature concatenation, as shown in Fig.5. Instead of copying features directly, these feature maps are fed

> REPLACE THIS LINE WITH YOUR PAPER IDENTIFICATION NUMBER (DOUBLE-CLICK HERE TO EDIT) <       7in the first DBB of DUs to exploit more contextual information. As shown in Fig.6(b), the hierarchical information is combined and computed through two cascaded convolution layers for effective feature collection and reconstruction. Moreover, an up-sampler with transpose convolution is applied to improve resolution, reducing channels and generating the output feature map with the resolution of $(2H \times 2W \times C_{k+1})$.

*2) Multi-scale Feature Aggregation Block*

Based on well-designed EUs and DUs, our proposed CPFNet encourages comprehensive feature exploitation and provides smooth information flow. The feature map computed from EUs contains complex contextual features and high-level semantic information. However, the diversity of aggregated features is still limited in our dual-scale feature extraction strategy applied in DBB. A multi-scale approach with more dilation rates should be employed for further feature collection, representation, and summarization for better semantic understanding parsing. Therefore, we proposed the Multi-scale Feature Aggregation Block (MFAB) for comprehensive analysis of multi-level multi-scale semantic features, as shown in Fig.6(c).

The MFAB is mainly composed of densely connected Multi-scale Aggregation Nodes (MAN) to achieve better information recycling. The MAN is also an upgraded version of the standard bottleneck block, where the second convolution layer is replaced by a multi-pathway residual structure. As shown in Fig. 5 (e). There are three pathways with different convolution dilation rates $(d_1, d_2, d_3)$ in MAN to obtain different feature maps which are added for fusion rather than being concatenated. Therefore, the multi-scale features are sufficiently computed and aggregated, providing abundant information.

The input feature map fed into MFAB is firstly introduced to three densely connected MANs, which achieves multi-scale information synthesis, as shown in Fig.5(c). Noted that the concatenation operation may increase the memory cost severely, the connected feature maps are compressed with a convolution layer to reduce channels. After the densely connected structure, the hierarchical feature map is obtained and the abundant semantic information is fed to the decoder module to construct the labels, achieving better face parsing performance.

*3) Loss Function Settings*

For the 2D semantic segmentation network, the weighted 2D cross-entropy loss function is employed in our study, given by:

$$\mathcal{L}_{I_i} = \frac{1}{HW}\sum_{x=1}^{H}\sum_{y=1}^{W}\left(-\sum_{l=1}^{N_l} w_l P_l^{(I_i,x,y)} \log Q_l^{(I_i,x,y)}\right) \quad (8)$$

where the resolution of image $I_i$ is $H \times W$.

From Eq. (8), it's noticed that the label weight affects the loss function directly. In our study, the label weights are set referring to the area of different labels, using the function:

$$w_l = \ln\left(1 + \frac{1}{N_I}\sum_{i=1}^{N_I}\frac{H \times W}{N_{(P_l=l)}^{(i)}}\right) \quad (9)$$

In the equation, the dataset contains $N_I$ images with the resolution of $(H \times W)$, and $N_{(P_l=l)}^{(i)}$ represents the number of pixels with label $l$ in the $i^{th}$ image. There may be great difference between $N_{(P_l=l)}$ of different label $l$, so a logarithm operation is applied to the weight calculation to control the gap between different label weights.

IV. EXPERIMENTS

To demonstrate the performance of our proposed 3D face parsing method, comprehensive experiments are carried out for evaluation. In this section, we first introduce details of our experimental settings. Then, we compare our CPFNet and 3D face parsing method with some state-of-the-art 2D semantic segmentation networks and 3D segmentation methods, respectively. Besides, we evaluate our method in various aspects to study the effect of different elements.

*A. Experimental Settings*

*1) Implementation Details*

Our CPFNet is trained from sketch with Adam optimizer with betas of (0.9,0.999) and weight decay of 0.0001. Moreover, exponential learning rate schedule with a warm-up strategy is employed, which is expressed as:

$$L = L_0 \times \left(1 - \frac{current\_iter}{maximum}\right)^{power} \quad (10)$$

where $L$ is the current learning rate, $L_0$ is the initial learning rate, and is set to $5 \times 10^{-4}$. The parameter called $power$ controls the speed of learning rate decay, which is set to 0.9. For the warm-up strategy, the warm-up factor is set to 0.3333, and the warm-up iteration is set to 500. The dropout strategy is employed in our network with the dropout rates in the range of 0 to 0.15.

*2) Datasets*

**BJTU-3D dataset [52]**: Portions of the research in this paper use the BJTU-3D Face Database collected under the joint sponsor of National Natural Science Foundation of China, Beijing Natural Science Foundation Program, Beijing Science and Educational Committee Program. BJTU-3D face dataset contains 500 3D face data including 250 faces of male and female each, ranging from 16 to 49 years old. All 3D faces are sampled with neutral expression and no occlusion situation such as hair or glasses is involved. The raw data in BJTU-3D dataset contains vertex coordinates, texture information, and grid data. In our study, the 3D face data is divided into training set, validation set, and testing set in the 4:1:1.

**CASIA-3D dataset [53]**: Portions of the research in this paper use the CASIA-3D FaceV1 collected by the Chinese Academy of Sciences Institute of Automation (CASIA). CASIA-3D FaceV1 dataset was sampled in 2004 using the on-contrast 3D digitizer. The dataset contains 3D face data from 123 people under various illumination. Compared with BJTU-3D dataset, CASIA-3D dataset was captured under different illumination, so that it is used to verify the advantages of our 3D face parsing against other 2D methods.

**UoY dataset [54]**: The 3D face dataset was constructed by the University of York and contains over 300 subjects scanned by Stereo vision 3D camera. Every 3D face data includes an OBJ data attached with a 2D image to record the texture information. Compared with the two datasets above, the UoY dataset represents texture information with gray value rather than RGB, and the resolution of UoY dataset is lower. The UoY dataset will further verify the robustness of our method.

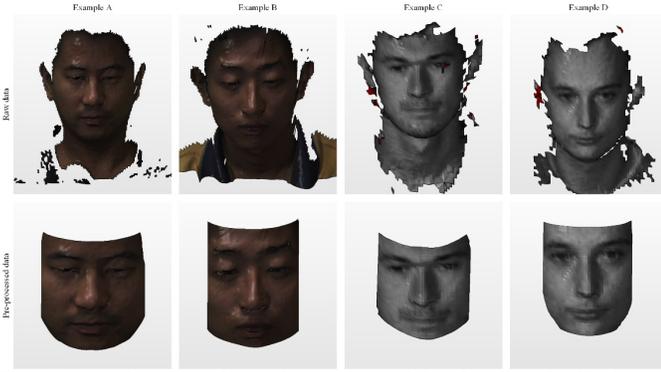

Fig. 7. Examples of raw data and pre-processed data from CASIA-3D dataset and UoY dataset

*3) Experimental Platform*

The experiment is conducted on a standard PC with Intel i5 9400F 2.90GHz and Nvidia GeForce RTX 2080 Ti. Because of the memory limitation, the batch size for training is 8, and the test batch size is 1 for all models.

The 3D face data pre-processing (described in the next sub-section) is based on Netfabb 2018. The face parameterization and the re-mapping process are coded and evaluated with Matlab R2020a, and the 2D semantic segmentation network is accomplished based on Pytorch.

*4) 3D Face Data Pre-processing*

In this paper, our method is evaluated based on public datasets as mentioned. Unfortunately, the 3D raw data in these datasets is captured incompletely in some non-face regions, such as neck and ears, and such areas are useless in face parsing study, so the raw data is not appropriate for 3D face parsing directly. Therefore, we remove the useless parts of the 3D raw data and maintain the critical part of the face containing the main facial components of eyebrows, eyes, nose, and mouth. Moreover, the 3D face data should be simply-connected open surface to meet the requirement of our parameterization, while there are still some holes in these 3D face data, especially in regions such as eyes. Therefore, the hole-filling process is applied before the face parameterization algorithm.

After pre-processing, the 3D face data is in the form as follows:

$$Vertex\ n\ x\ y\ z\ r\ g\ b$$
$$Face\ m\ v1\ v2\ v3$$

where $n$ and $m$ are the indexes of vertexes and triangular patches, respectively. Every vertex has a coordinate of $(x, y, z)$ with the color feature of $(r, g, b)$, and every triangular patch (*i.e.*, Face) is composed of three vertexes of $v1$, $v2$, and $v3$.

*5) Parameterization Settings and 3D Method settings*

The number of vertices and triangle patches is free for our face parameterization algorithm. The 2D face images generated from parameterization are resized to $256 \times 256$, and all channels of the image are normalized before fed into CPFNet.

However, the number of input points needs to be the same during experiments of 3D point cloud methods, so point cloud resampling is applied during some experiments. Noted that the 3D data resolutions in the three datasets are different, different numbers of points are resampled: 8192 points for BJTU-3D dataset, while 4096 points for CASIA-3D dataset and UoY dataset.

*6) Evaluation Metrics*

The $Mean\ IoU$ ($MIoU$) is used as the quantitative metric for the evaluation of face parsing performance. Noted that both 2D face image and 3D mesh data are involved in this study, the labels should be measured by pixel number for 2D image segmentation and area for 3D mesh data. Therefore, two types of MIoU are defined for two measurements respectively.

**$Mean\ IoU$ ($MIoU$):** Average ratio of correctly predicted regions with label $l$ over the union set of three parts: correctly predicted regions with label $l$, regions with true label $l$ but predicted as label $i (i \neq l)$, and regions with real label $i (i \neq l)$ but predicted as label $l$, *i.e.*,

$$MIoU = \frac{1}{N-1} \sum_{l=2}^{N} \frac{Pred_{l,l}}{\sum_i Pred_{l,i} + \sum_i Pred_{i,l} - Pred_{l,l}} \quad (11)$$

In this study, $N$ is 5, but the background is only involved in measuring the union set of areas. The IoU of the background is not used in MIoU to measure the performance of main facial component segmentation, so $i$ ranges from 1 to 5 while $l$ starts from 2. $Pred_{l,i}$ illustrates the region that is labeled as $l$ and predicted as $i$, which is different in two types of MIoU. So, the different expressions can be given as:

$$MIoU_{2D} = \frac{1}{N-1} \sum_{l=2}^{N} \frac{N_{l,l}^{(Pixel)}}{\sum_i N_{l,i}^{(Pixel)} + \sum_i N_{i,l}^{(Pixel)} - N_{l,l}^{(Pixel)}} \quad (12)$$

for 2D images where $N^{(Pixel)}$ represents the number of pixels with specific labels, and

$$MIoU_{3D} = \frac{1}{N-1} \sum_{l=2}^{N} \frac{\sum_l S_{l,l}}{\sum_i S_{l,i} + \sum_i S_{i,l} - S_{l,l}} \quad (13)$$

for 3D surface where $S$ represents the area of a triangle patch with the specific label.

*B. Study of Network Elements*

In this subsection, we compare different the performance of models with different structures to verify the effectiveness of our proposed modules.

*1) Comparison of DBB:*

As a critical structure of our encoder-decoder network, the design of DBB has a great influence on feature exploitation and information analysis, which impacts the performance directly. To verify the effectiveness of our dual-scale strategy, we compare the performance between basic convolution layers, standard bottleneck structure, and DBB with different dilation rate settings. As results shown in Table I, our dual-scale strategy improves the face parsing accuracy compared with the standard bottleneck block and the double convolution layers. Meanwhile, dilation rates of $(1, 2)$ provide the best performance.

*2) Comparison of EU:*

As a main module of CPFNet, the EU collects multi-scale features for face information understanding. As the result of models with different EUs listed in Table II, the performance improves as the number of DBB increases from 1 to 3, with the mIoU rising from 80.64% to 82.37%. However, when more DBBs are applied, the improvement becomes tiny due to the increasing network depth. Therefore, it is proved that EU with

TABLE I
EVALUATION OF DUAL-SCALE STRATEGY

| Activation | Dilation Rates | Mean IoU (%) |
|---|---|---|
| Double Convolution Layers | --- | 79.41 |
| Standard Bottleneck Block | --- | 80.39 |
| Dual-scale Bottleneck Block | (1, 1) | 81.26 |
| | (1, 2) | 82.35 |
| | (1, 3) | 81.93 |
| | (1, 4) | 81.78 |

All the other designs in this comparison experiment are the same as the network structure mentioned in Chapter III.

TABLE II
EVALUATION OF DUAL-SCALE BOTTLENECK BLOCK IN ENCODER UNITS

| Number of DBB | Mean IoU (%) | Number of Parameters |
|---|---|---|
| 1 | 80.64 | 2.06M |
| 2 | 81.65 | 2.33M |
| 3 | 82.35 | 2.60M |
| 4 | 82.21 | 2.87M |
| 5 | 82.40 | 3.14M |

All the other designs in this comparison experiment are the same as the network structure mentioned in Chapter III.

TABLE III
EVALUATION OF MULTI-SCALE AGGREGATION STRATEGY

| Number of Pathways | Mean IoU (%) | Number of Parameters |
|---|---|---|
| 1 | 80.95 | 2.54M |
| 2 | 81.67 | 2.57M |
| 3 | 82.35 | 2.60M |
| 4 | 82.13 | 2.63M |
| 5 | 82.31 | 2.66M |

All the other designs in this comparison experiment are the same as the network structure mentioned in Chapter III.

TABLE IV
EVALUATION OF MFAB

| Number of MAN | Mean IoU (%) | Number of Parameters |
|---|---|---|
| 0 | 81.50 | 2.17M |
| 1 | 81.77 | 2.28M |
| 2 | 82.35 | 2.60M |
| 3 | 82.29 | 3.13M |
| 4 | 82.36 | 3.87M |

All the other designs in this comparison experiment are the same as the network structure mentioned in Chapter III.

cascaded DBBs ameliorates the parameterized face parsing, while the increasing parameters may cause inefficiency.

*3) Comparison of MFAB:*

As the module for multi-scale semantic feature computation, MAFB achieves abundant information analysis and feature fusion with Multi-scale Aggregation Nodes (MAN). In this experiment, we use networks employed with different MAN and MFAB respectively, to study the effect of the number of multi-scale pathways in MAN and the design of dense structure in MFAB for better feature fusion. The results are listed in Table III and Table IV respectively. As listed in Table III, the best result is obtained with 3 multi-scale pathways for feature exploitation of different receptive fields. As for the number of MAN in our densely connected structure, the double-MAN design shows the best performance among all groups, obtaining the best mIoU with an acceptable amount of network parameters.

From the results displayed above, it can be proved that our proposed modules are effective and promote better learning of semantic information in 2D disk-like images, providing the better performance of facial component segmentation.

*C. Comparison with State-of-the-art 2D Network*

To evaluate the effectiveness of CPFNet, we compared our method with other state-of-the-art 2D semantic segmentation networks. Referring to the quantitative results, our CPFNet performers better on all three datasets.

*1) Results on BJTU-Dataset*

The comparison results on BJTU-3D dataset are listed in Table V, with a few examples shown in Fig.7. Among all networks, our CPFNet achieves the highest scores on the BJTU-3D datasets. In the experiment, widely applied networks such as UNet and DeepLab get lower scores than our method due to error segmentation. With distorted face images as input, deep learning techniques such as atrous convolution, ASPP, and attention, and increasing model parameters cannot bring obvious improvement to face parsing performance. To prove the outstanding efficiency, CPFNet is compared with lightweight networks with the elaborate structure. As presented in Table V, our CPFNet also achieves the best results. Therefore, it is essential to design specific network structures for comprehensive information collection of the data. To further demonstrate the performance of CPFNet on face parsing, state-of-the-art 2D face parsing networks for 2D images are also involved in this experiment. Although the CPFNet is slightly inferior to EHANet18 with boundary awareness on nose segmentation, our method still holistically outperforms face parsing networks.

Based on comparison results listed above, it is proved that our proposed CPFNet for parameterized face parsing is both effective and efficient.

*2) Results on CASIA-3D Dataset and UoY Dataset*

As mentioned before, CASIA-3D and UoY datasets contain images with different illumination and low-resolution gray images respectively Comparison with state-of-the-art 2D networks based on these two datasets are applied for further evaluation of robustness and generalization.

Experiment results on both two datasets are listed in Table VI. For the experiments on CASIA-3D dataset, common conventional networks and lightweight networks obtain acceptable face parsing results, while the CPFNet achieves the best MIoU of 81.75%. Compared with both BJTU and CASIA-3D datasets, the 3D face data in UoY dataset only contains gray value and the resolution is lower, which causes the performance to drop greatly. However, CPFNet still achieves the best mIoU of 70.13%, as listed in Table. V. Therefore, experimental results on these two datasets further demonstrate the effectiveness and robustness of our CPFNet.

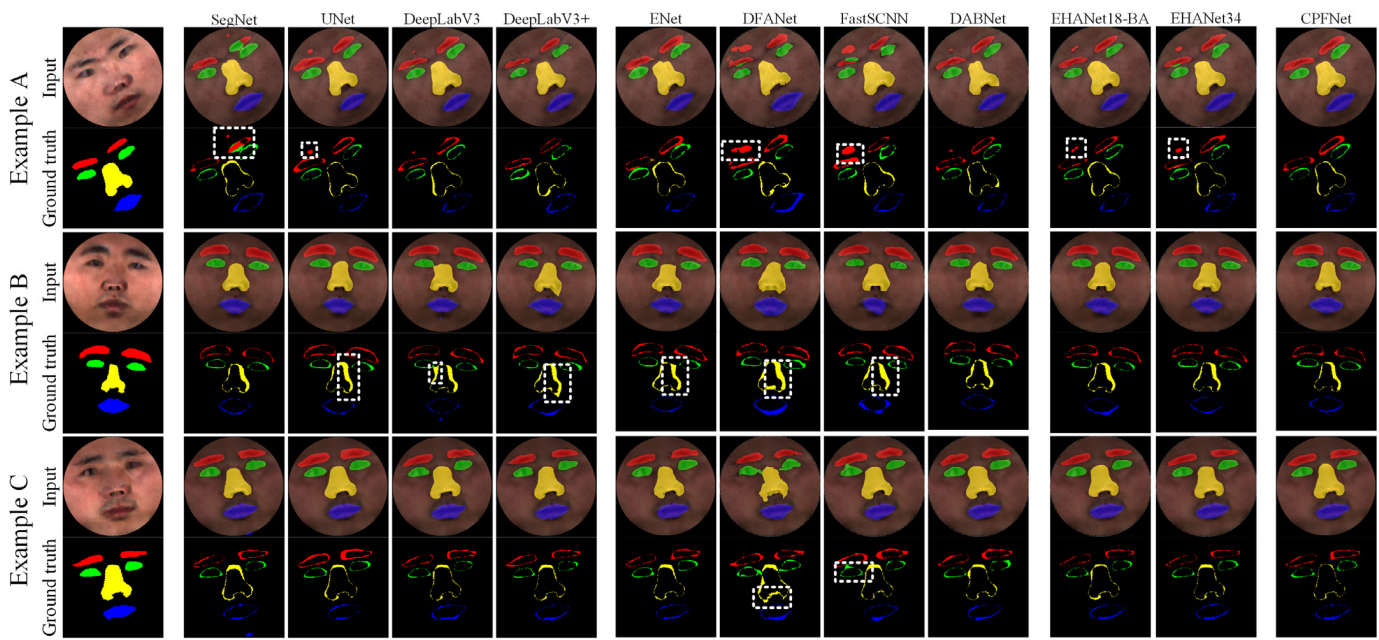

Fig. 8. Comparison with other state-of-the-art networks. Both predicted labels and the difference between prediction and ground truth are displayed in the figure.

TABLE V
COMPARISONS WITH STATE-OF-THE-ART 2D NETWORKS ON BJTU-3D DATASET

| Model | EYEBROW (%) | Eye (%) | Nose (%) | Mouth (%) | MIoU (%) | Parameter |
|---|---|---|---|---|---|---|
| SegNet [55] | 75.83 | 80.07 | 79.45 | 83.71 | 79.76 | 34.76M |
| UNet [56] | 75.88 | 80.46 | 82.26 | 84.85 | 80.86 | 14.16M |
| DeepLabV3 (Res50) [57] | 75.53 | 78.97 | 83.25 | 85.16 | 80.13 | 39.64M |
| DeepLabV3+ (Res50) [58] | 76.91 | 78.46 | 81.98 | 85.01 | 80.59 | 39.76M |
| DANet [59] | 75.29 | 77.93 | 79.35 | 83.31 | 80.07 | 68.48M |
| ENet [60] | 75.79 | 79.85 | 80.35 | 84.18 | 80.05 | 0.36M |
| ERFNet [61] | 75.18 | 78.55 | 80.14 | 85.47 | 79.84 | 2.06M |
| ESNet [62] | 76.03 | 81.07 | 82.49 | 85.25 | 81.20 | 1.66M |
| DABNet [63] | 73.79 | 77.70 | 81.27 | 83.73 | 79.17 | 0.75M |
| DFANet [64] | 63.10 | 71.09 | 77.35 | 72.88 | 71.10 | 7.73M |
| FastSCNN [65] | 64.85 | 71.41 | 78.29 | 76.41 | 72.74 | 1.14M |
| CGNet [66] | 75.47 | 79.84 | 82.13 | 83.80 | 79.31 | 0.49M |
| EHANet18 [11] | 74.36 | 78.78 | 83.13 | 84.61 | 79.20 | 11.73M |
| EHANet18 (Boundary Awareness) [11] | 75.86 | 78.45 | **83.43** | 84.80 | 80.63 | 11.73M |
| EHANet34 [11] | 74.76 | 78.44 | 82.99 | 84.61 | 80.20 | 21.83M |
| EHANet34 (Boundary Awareness) [11] | 74.60 | 78.83 | 83.01 | 85.28 | 80.43 | 21.83M |
| CPFNet | **78.23** | **82.83** | 82.55 | **86.12** | **82.35** | 2.60M |

### D. Comparison with Other 3D Methods

The 2D face image semantic segmentation is the intermediate process in the 3D face parsing, so it cannot comprehensively and completely evaluate the improvement of 3D face parsing method with the "3D-2D-3D" strategy. Therefore, it is necessary to compare our method with other 3D face parsing approaches. Pointwise MLP methods, point cloud convolution networks, graph convolution methods, and a voxelization method are involved in the experiments.

*1) Results on BJTU -Dataset*

The comparison results are presented in Table VII, and several 3D face parsing results based on different methods are shown in Fig.8. In the comparison with other methods, our method achieves the highest scores on the BJTU dataset. All methods with point-cloud-based networks cannot provide

> REPLACE THIS LINE WITH YOUR PAPER IDENTIFICATION NUMBER (DOUBLE-CLICK HERE TO EDIT) <    11TABLE VI
COMPARISON WITH STATE-OF-THE-ART 2D NETWORKS ON CASIA-3D DATASET AND UOY DATASET

| Model | MIoU of CASIA-3D Dataset (%) | MIoU of UoY Dataset (%) |
|---|---|---|
| SegNet [55] | 78.93 | 67.31 |
| UNet [56] | 80.04 | 67.86 |
| DeepLabV3 (ResNet50) [57] | 80.91 | 67.87 |
| DeepLabV3 (MobileNet) [57] | 74.25 | 63.92 |
| DeepLabV3+ (ResNet50) [58] | 81.13 | 68.14 |
| DeepLabV3+ (MobileNet) [58] | 76.43 | 64.71 |
| DANet [59] | 79.51 | 67.08 |
| ENet [60] | 79.02 | 67.30 |
| ERFNet [61] | 80.72 | 68.23 |
| ESNet [62] | 81.39 | 69.19 |
| DABNet [63] | 79.83 | 66.13 |
| DFANet [64] | 74.98 | 66.46 |
| FastSCNN [65] | 75.60 | 61.27 |
| CGNet [66] | 81.14 | 65.45 |
| EHANet18 [11] | 81.20 | 68.27 |
| EHANet18 (Boundary Awareness) [11] | 81.49 | 68.80 |
| EHANet34 [11] | 81.08 | 68.40 |
| EHANet34 (Boundary Awareness) [11] | 81.42 | 68.54 |
| **CPFNet** | **81.75** | **70.13** |

accurate segmentation because of the smooth surface which causes unclear prediction of the region edge.

Graph convolution networks DGCNN provides some good 3D face parsing results with the mIoU of 75.10%, showing the effectiveness of EdgeConv technology and the dynamic graph strategy. In addition, compared with the other 3D segmentation methods, the voxelization method achieves acceptable results and better nose segmentation than our method. However, better global performance is presented by our method, and the voxelization method has the lowest resolution due to the 3D grid projection.

*2) Results on CASIA-3D Dataset*

The performance on CASIA-3D dataset is slightly worse than that on BJTU dataset due to its lower resolution and various illumination, as presented in Table VIII. In general, the low resolution and various illumination limit the spatial feature synthesis and texture information analysis of the network, respectively. However, as shown in Table VII, our face parsing method still achieves acceptable results and the best scores among all 3D segmentation methods.

## V. CONCLUSION

In this paper, we propose a 3D face parsing method via face parameterization and a 2D convolutional network to extract the pixel-wise level semantic constituents (e.g., mouth, eyes and nose). The "3D-2D-3D" strategy is introduced in our method to compute the parsing of 3D mesh data. The 3D data is first transformed to a 2D disk domain through the face parameterization algorithm to compute the distorted disk-like 2D face image with both textural and spatial information. The convolution neural network called CPFNet is proposed for semantic segmentation of parameterized face, and the semantic labels are inversely re-mapped to the 3D face data, which finally achieves the 3D face parsing. The proposed method is evaluated on data from different datasets after pre-processing. Compared with the other state-of-the-art 2D semantic segmentation network and 3D methods, our CPFNet presents better parsing results on parameterized 2D face images, and our 3D face parsing method considerably improves the 3D face parsing performance. Experiments proved that modules of CPFNet improve feature extraction and information aggregation. Therefore, our 3D face parsing method via face parameterization and 2D network shows immense potential on application in advanced 3D face technologies.

## VI. REFERENCE

[1] D. Zhang, L. Lin, T. Chen, X. Wu, W. Tan, and E. Izquierdo, "Content-Adaptive Sketch Portrait Generation by Decompositional Representation Learning," *IEEE Transactions on Image Processing*, vol. 26, no. 1, pp. 328–339, Jan. 2017, doi: 10.1109/tip.2016.2623485.
[2] X. Ou, S. Liu, X. Cao, and H. Ling, "Beauty emakeup: A deep makeup transfer system," in *Proceedings of the 24th ACM international conference on Multimedia*, Oct. 2016, pp. 701–702.
[3] K. Khan, M. Attique, R. U. Khan, I. Syed, and T.-S. Chung, "A Multi-Task Framework for Facial Attributes Classification through End-to-End Face Parsing and Deep Convolutional Neural Networks," *Sensors*, vol. 20, no. 2, p. 328, Jan. 2020, doi: 10.3390/s20020328.
[4] X. Jin *et al.*, "Face Illumination Transfer and Swapping via Dense Landmark and Semantic Parsing," *IEEE Sensors Journal*, p. , 2020, doi: 10.1109/jsen.2020.3025918.
[5] Z. Shen, W.-S. Lai, T. Xu, J. Kautz, and M.-H. Yang, "Deep semantic face deblurring," in *Proceedings of the IEEE Conference on Computer Vision and Pattern Recognition*, 2018, pp. 8260--8269.
[6] T. Li *et al.*, "Beautygan: Instance-level facial makeup transfer with deep generative adversarial network," in *Proceedings of the 26th ACM international conference on Multimedia*, 2018, pp. 645–653.
[7] A. S. Jackson, M. Valster, and G. Tzimiropoulos, "A CNN cascade for landmark guided semantic part segmentation," in *European Conference on Computer Vision*, Sep. 2016, pp. 143–155.
[8] Z. Wei, S. Liu, Y. Sun, and H. Ling, "Accurate Facial Image Parsing at Real-Time Speed," *IEEE Transactions on Image Processing*, vol. 28, no. 9, pp. 4659–4670, Sep. 2019, doi: 10.1109/tip.2019.2909652.
[9] J. Warrell and S. J. Prince, "Labelfaces: Parsing facial features by multiclass labeling with an epitome prior," in *2009 16th IEEE international conference on image processing (ICIP)*, 2009, pp. 2481--2484.
[10] Z. Wei, Y. Sun, and J. Wang, et al. "Learning adaptive receptive fields for deep image parsing network," *Proceedings of the IEEE conference on computer vision and pattern recognition*. pp. 2434-2442, 2017.
[11] L. Luo, D. Xue, and X. Feng, "EHANet: An Effective Hierarchical Aggregation Network for Face Parsing," *Applied Sciences*, vol. 10, no. 9, p. 3135, Apr. 2020, doi: 10.3390/app10093135.
[12] J. Lin, H. Yang, D. Chen, M. Zeng, F. Wen, and L. Yuan, "Face parsing with roi tanh-warping," in *Proceedings of the IEEE/CVF Conference on Computer Vision and Pattern Recognition*, 2019, pp. 5654--5663.
[13] X. Li, D. Yang, Y. Wang, W. Zhang, F. Li, and W. Zhang, "TCMINet: Face Parsing for Traditional Chinese Medicine Inspection via a Hybrid Neural Network with Context Aggregation," *IEEE Access*, vol. 8, pp. 93069–93082, May 2020, doi: 10.1109/access.2020.2995202.
[14] Y. Yu, K. A. F. Mora, and J.-M. Odobez, "Robust and accurate 3d head pose estimation through 3dmm and online head model reconstruction," in *2017 12th ieee international conference on automatic face \& gesture recognition (fg 2017)*, 2017, pp. 711–718.

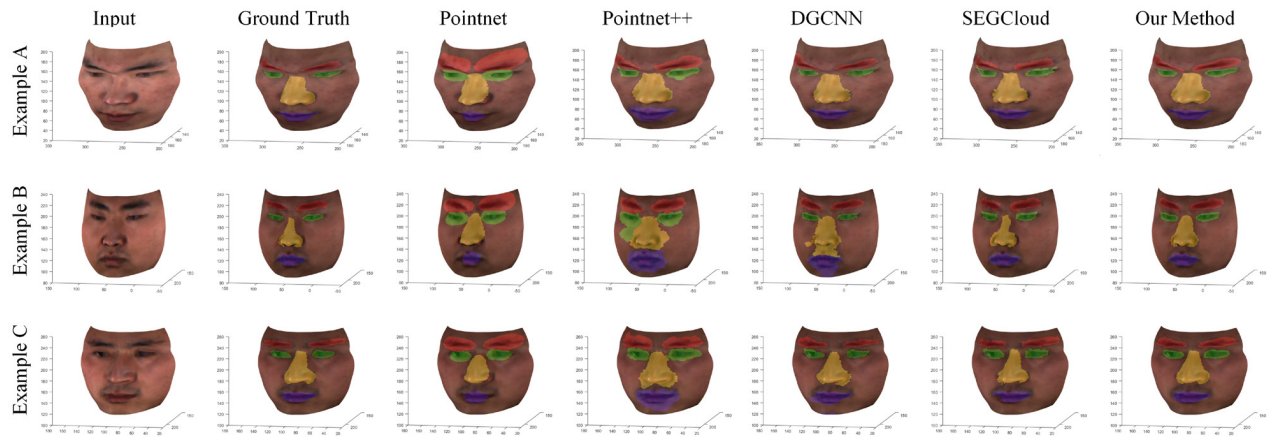

Fig. 9. Comparison with other state-of-the-art 3D methods, including point cloud methods and voxelization methods.

TABLE VII
COMPARISON WITH STATE-OF-THE-ART 3D METHODS ON BJTU-3D DATASET

|  | MODEL | EYEBROW (%) | Eye (%) | Nose (%) | Mouth (%) | MIoU (%) |
|---|---|---|---|---|---|---|
| Point Cloud Network Method | PointNet [20] | 45.97 | 50.35 | 56.55 | 47.67 | 50.13 |
|  | PointNet++ [21] | 57.24 | 54.62 | 67.45 | 59.00 | 59.58 |
|  | KP-Conv [22] | 63.82 | 75.94 | 78.53 | 70.26 | 72.14 |
|  | Point-CNN [47] | 61.33 | 76.02 | 74.86 | 70.05 | 70.48 |
| GCN-based Method | GACNet [23] | 60.25 | 58.58 | 57.83 | 69.69 | 61.59 |
|  | DGCNN [24] | 67.51 | 79.24 | 81.71 | 71.94 | 75.10 |
| Voxelization Method | Seg-Cloud [18] | 75.57 | 71.70 | **83.75** | 79.35 | 77.59 |
| Our Method | 3D-2D-3D | **78.23** | **82.83** | 82.55 | **86.12** | **82.10** |

TABLE VIII
COMPARISON WITH STATE-OF-THE-ART 3D METHODS ON CASIA-3D DATASET

|  | MODEL | MIoU ON CASIA-3D DATASET (%) |
|---|---|---|
| Point Cloud Network Method | PointNet [20] | 49.71 |
|  | PointNet++ [21] | 52.84 |
|  | KP-Conv [22] | 66.30 |
|  | Point-CNN [47] | 65.08 |
| GCN-based Method | GACNet [23] | 59.73 |
|  | DGCNN [24] | 68.94 |
| Voxelization Method | Seg-Cloud [18] | 74.06 |
| Our Method | 3D-2D-3D | **80.14** |

[15] P. Paysan, R. Knothe, B. Amberg, S. Romdhani, and T. Vetter, "A 3D face model for pose and illumination invariant face recognition," in *2009 sixth IEEE international conference on advanced video and signal based surveillance*, 2009, pp. 296--301.
[16] Z. Wang, "Robust three-dimensional face reconstruction by one-shot structured light line pattern," *Optics and Lasers in Engineering*, vol. 124, p. 105798, Jan. 2020, doi: 10.1016/j.optlaseng.2019.105798.
[17] J. Kim, S. Park, S. Kim, and S. Lee, "Registration method between ToF and color cameras for face recognition," in *2011 6th IEEE Conference on Industrial Electronics and Applications*, 2011, pp. 1977--1980.
[18] L. Tchapmi, C. Choy, I. Armeni, J. Gwak, and S. Savarese, "Segcloud: Semantic segmentation of 3d point clouds," in *2017 international conference on 3D vision (3DV)*, 2017, pp. 537–547.
[19] H.-Y. Meng, L. Gao, Y.-K. Lai, and D. Manocha, "Vv-net: Voxel vae net with group convolutions for point cloud segmentation," in *Proceedings of the IEEE/CVF International Conference on Computer Vision*, 2019, pp. 8500—8508.
[20] C. R. Qi, H. Su, K. Mo, and L. J. Guibas, "Pointnet: Deep learning on point sets for 3d classification and segmentation," in *Proceedings of the IEEE conference on computer vision and pattern recognition*, 2017, pp. 652–660.
[21] C. R. Qi, L. Yi, H. Su, and L. J. Guibas, "Pointnet++: Deep hierarchical feature learning on point sets in a metric space," *arXiv preprint arXiv:1706.02413*, 2017.
[22] H. Thomas, C. R. Qi, J.-E. Deschaud, B. Marcotegui, F. Goulette, and L. J. Guibas, "Kpconv: Flexible and deformable convolution for point clouds," in *Proceedings of the IEEE/CVF International Conference on Computer Vision*, 2019, pp. 6411–6420.
[23] Y. Wang, Y. Sun, Z. Liu, S. E. Sarma, M. M. Bronstein, and J. M. Solomon, "Dynamic Graph CNN for Learning on Point Clouds," *ACM Transactions on Graphics*, vol. 38, no. 5, pp. 1–12, Nov. 2019, doi: 10.1145/3326362.
[24] L. Wang, Y. Huang, Y. Hou, S. Zhang, and J. Shan, "Graph attention convolution for point cloud semantic segmentation," in *Proceedings of the IEEE/CVF Conference on Computer Vision and Pattern Recognition*, 2019, pp. 10296–10305.
[25] H. Su, S. Maji, E. Kalogerakis, and E. Learned-Miller, "Multi-view convolutional neural networks for 3d shape recognition," in *Proceedings of the IEEE international conference on computer vision*, 2015, pp. 945--953.
[26] J. L. Felix, M. Danelljan, P. Tosteberg, G. Bhat, K. Fahad Shahbaz, and F. Michael, "Deep projective 3D semantic segmentation," in *International Conference on Computer Analysis of Images and Patterns*, 2017, pp. 95--107.
[27] X. Qi, R. Liao, J. Jia, S. Fidler, and R. Urtasun, "3d graph neural networks for rgbd semantic segmentation," in *Proceedings of the IEEE International Conference on Computer Vision*, 2017, pp. 5199--5208.

[28] X. Hu, K. Yang, L. Fei, and K. Wang, "Acnet: Attention based network to exploit complementary features for rgbd semantic segmentation," in *2019 IEEE International Conference on Image Processing (ICIP)*, 2019, pp. 1440--1444.

[29] X. Gu and S.-T. Yau, "Global conformal surface parameterization," in *Proceedings of the 2003 Eurographics/ACM SIGGRAPH symposium on Geometry processing*, 2003, pp. 127–137.

[30] X. Gu, Y. Wang, T. F. Chan, P. M. Thompson, and S.-T. Yau, "Genus Zero Surface Conformal Mapping and Its Application to Brain Surface Mapping," *IEEE Transactions on Medical Imaging*, vol. 23, no. 8, pp. 949–958, Aug. 2004, doi: 10.1109/tmi.2004.831226.

[31] L. Kharevych, B. Springborn, and P. Schröder, "Discrete conformal mappings via circle patterns," *ACM Transactions on Graphics*, vol. 25, no. 2, pp. 412–438, Apr. 2006, doi: 10.1145/1138450.1138461.

[32] M. Jin, Y. Wang, S-T. Yau, and X. Gu, "Optimal global conformal surface parameterization," in *IEEE Visualization 2004*, 2004, pp. 267--274.

[33] Y. Wang *et al.*, "Brain Surface Conformal Parameterization Using Riemann Surface Structure," *IEEE Transactions on Medical Imaging*, vol. 26, no. 6, pp. 853–865, Jun. 2007, doi: 10.1109/tmi.2007.895464.

[34] Y. Wang *et al.*, "Brain Surface Conformal Parameterization With the Ricci Flow," *IEEE Transactions on Medical Imaging*, vol. 31, no. 2, pp. 251–264, Feb. 2012, doi: 10.1109/tmi.2011.2168233.

[35] M. Jin, J. Kim, F. Luo, and X. Gu, "Discrete Surface Ricci Flow," *IEEE Transactions on Visualization and Computer Graphics*, vol. 14, no. 5, pp. 1030–1043, Sep. 2008, doi: 10.1109/tvcg.2008.57.

[36] F. Luo, "Combinatorial Yamabe flow on surfaces," *Communications in Contemporary Mathematics*, vol. 6, no. 5, pp. 765--780, 2004.

[37] P. T. Choi and L. M. Lui, "Fast disk conformal parameterization of simply-connected open surfaces," *Journal of Scientific Computing*, vol. 65, no. 3, pp. 1065--1090, 2015.

[38] G. P. T. Choi, "Efficient Conformal Parameterization of Multiply-Connected Surfaces Using Quasi-Conformal Theory," *Journal of Scientific Computing*, vol. 87, no. 3, Apr. 2021, doi: 10.1007/s10915-021-01479-y.

[39] B. M. Smith, L. Zhang, J. Brandt, Z. Lin, and J. Yang, "Exemplar-based face parsing," in *Proceedings of the IEEE conference on computer vision and pattern recognition*, 2013, pp. 3484--3491.

[40] P. Luo, X. Wang, and X. Tang, "Hierarchical face parsing via deep learning," *2012 IEEE Conference on Computer Vision and Pattern Recognition*, pp. 2480--2487, 2012.

[41] S. Liu, J. Yang, C. Huang, and M.-H. Yang, "Multi-objective convolutional learning for face labeling," in *Proceedings of the IEEE Conference on Computer Vision and Pattern Recognition*, 2015, pp. 3451--3459.

[42] J. Long, E. Shelhamer, and T. Darrell, "Fully Convolutional Networks for Semantic Segmentation," in Proceesings of the IEEE conference on computer vision and pattern recognition, Jun. 2015, pp. 3431-3440.

[43] T. Guo et al., "Residual encoder decoder network and adaptive prior for face parsing," in *Proceedings of the AAAI Conference on Artificial Intelligence*, 2018, vol. 32, no. 1.

[44] Te G, Hu W, Liu Y, et al. AGRNet: Adaptive Graph Representation Learning and Reasoning for Face Parsing[J]. *IEEE Transactions on Image Processing*, 2021.

[45] W. Chu, W.-C. Hung, Y.-H. Tsai, D. Cai, and M.-H. Yang, "Weakly-supervised caricature face parsing through domain adaptation," in *2019 IEEE International Conference on Image Processing (ICIP)*, 2019, pp. 3282–3286.

[46] M. Tatarchenko, J. Park, V. Koltun, and Q.-Y. Zhou, "Tangent convolutions for dense prediction in 3d," in *Proceedings of the IEEE Conference on Computer Vision and Pattern Recognition*, 2018, pp. 3887--3896.

[47] Y. Li, R. Bu, M. Sun, W. Wu, X. Di, and B. Chen, "Pointcnn: Convolution on x-transformed points," *Advances in neural information processing systems*, vol. 31, pp. 820--830, 2018.

[48] R. Hanocka, A. Hertz, N. Fish, R. Giryes, S. Fleishman, and D. Cohen-Or, "MeshCNN," *ACM Transactions on Graphics*, vol. 38, no. 4, pp. 1–12, Jul. 2019, doi: 10.1145/3306346.3322959.

[49] A. Sinha, J. Bai, and K. Ramani, "Deep learning 3D shape surfaces using geometry images," in *European conference on computer vision*, 2016, pp. 223–240.

[50] S. Tulsiani, H. Su, L. J. Guibas, A. A. Efros, and J. Malik, "Learning shape abstractions by assembling volumetric primitives," in *Proceedings of the IEEE Conference on Computer Vision and Pattern Recognition*, 2017, pp. 2635–2643.

[51] Y. Gong, L. Wang, R. Guo, and S. Lazebnik, "Multi-scale orderless pooling of deep convolutional activation features," in *European conference on computer vision*, 2014, pp. 392–407.

[52] B. Yin, Y. Sun, C. Wang, and Y. Ge, "BJUT-3D large scale 3D face database and information processing," *Journal of Computer Research and Development*, vol. 46, no. 6, p. 1009, 2009.

[53] "CASIA-3D FaceV1." http://biometrics.idealtest.org/.

[54] T. Heseltine, N. Pears, and J. Austin, "Three Dimensional Face Recognition using Combinations of Surface Feature Map Subspace Components," *Image and Vision Computing*, vol. 26, no. 3, pp. 382–396, 2008.

[55] V. Badrinarayanan, A. Kendall, and R. Cipolla, "SegNet: A Deep Convolutional Encoder-Decoder Architecture for Image Segmentation," *IEEE Transactions on Pattern Analysis and Machine Intelligence*, vol. 39, no. 12, pp. 2481–2495, Dec. 2017, doi: 10.1109/tpami.2016.2644615.

[56] O. Ronneberger, P. Fischer, and T. Brox, "U-net: Convolutional networks for biomedical image segmentation," in *International Conference on Medical image computing and computer-assisted intervention*, 2015, pp. 234–241.

[57] Chen L C, Papandreou G, Schroff F, et al. Rethinking atrous convolution for semantic image segmentation[J]. *arXiv preprint arXiv:1706.05587*, 2017.

[58] L.-C. Chen, Y. Zhu, G. Papandreou, F. Schroff, and H. Adam, "Encoder-decoder with atrous separable convolution for semantic image segmentation," in *Proceedings of the European conference on computer vision (ECCV)*, 2018, pp. 801–818.

[59] J. Fu *et al.*, "Dual attention network for scene segmentation," in *Proceedings of the IEEE/CVF Conference on Computer Vision and Pattern Recognition*, 2019, pp. 3146–3154.

[60] A. Paszke, A. Chaurasia, S. Kim, and E. Culurciello, "Enet A deep neural network architecture for real-time semantic segmentation," *arXiv preprint arXiv:1606.02147*, 2016.

[61] E. Romera, J. M. Alvarez, L. M. Bergasa, and R. Arroyo, "ERFNet: Efficient Residual Factorized ConvNet for Real-Time Semantic Segmentation," *IEEE Transactions on Intelligent Transportation Systems*, vol. 19, no. 1, pp. 263–272, Jan. 2018, doi: 10.1109/tits.2017.2750080.

[62] H. Lyu, H. Fu, X. Hu, and L. Liu, "Esnet: Edge-Based Segmentation Network for Real-Time Semantic Segmentation in Traffic Scenes," in *2019 IEEE International Conference on Image Processing (ICIP)*, 2019, pp. 1855–1859.

[63] G. Li, I. Yun, J. Kim, and J. Kim, "Dabnet Depth-wise asymmetric bottleneck for real-time semantic segmentation," 2019.

[64] H. Li, P. Xiong, H. Fan, and J. Sun, "Dfanet: Deep feature aggregation for real-time semantic segmentation," in *Proceedings of the IEEE/CVF Conference on Computer Vision and Pattern Recognition*, 2019, pp. 9522–9531.

[65] R. P. Poudel, S. Liwicki, and R. Cipolla, "Fast-scnn: fast semantic segmentation network," *arXiv preprint arXiv:1902.04502*, 2019.

[66] T. Wu, S. Tang, R. Zhang, J. Cao, and Y. Zhang, "CGNet: A Light-Weight Context Guided Network for Semantic Segmentation," *IEEE Transactions on Image Processing*, vol. 30, pp. 1169–1179, 2021, doi: 10.1109/tip.2020.3042065.